\documentclass[11pt,a4paper]{article}
\usepackage[hyperref,acceptedWithA]{tacl2018v2} 
\usepackage{times,latexsym}
\usepackage{url}
\usepackage[T1]{fontenc}

\usepackage{xspace,mfirstuc,tabulary}

\usepackage{times}
\usepackage{latexsym}
\usepackage{amsmath}
\usepackage{multirow}
\usepackage{url}
\usepackage{makecell}
\usepackage{amsthm}
\usepackage{array,graphicx}
\usepackage{color,colortbl}
\usepackage{arydshln}
\definecolor{Gray}{gray}{0.93}

\usepackage{amsfonts}
\usepackage{amsmath}

\usepackage{xspace,mfirstuc,tabulary}

\newif\iftaclinstructions
\taclinstructionsfalse 
\iftaclinstructions
\renewcommand{\confidential}{}
\renewcommand{\anonsubtext}{(No author info supplied here, for consistency with
TACL-submission anonymization requirements)}
\newcommand{\instr}
\fi

\iftaclpubformat 

\else

\fi

\title{GILE: A Generalized Input-Label Embedding for Text Classification }

\author{Nikolaos Pappas  \ \ \ \ \ \ \ \ James Henderson\\
  Idiap Research Institute, Martigny 1920, Switzerland\\ 
  {\tt  \{nikolaos.pappas,james.henderson@idiap.ch\}} 
}

\date{}

\begin{document}

\maketitle

\begin{abstract} 
Neural text classification models typically treat output labels as categorical variables which lack description and semantics. This forces their parametrization to be dependent on the label set size, and, hence, they are unable to scale to large label sets and generalize to unseen ones. Existing joint input-label text models overcome these issues by exploiting label descriptions, but they are unable to capture complex label  relationships, have rigid parametrization, and their gains on unseen labels happen often at the expense of weak performance on the labels seen during training.
In this paper, we propose a new input-label model which generalizes over previous such models,  addresses their limitations, and does not compromise performance on  seen labels. The model consists of a joint non-linear input-label embedding with controllable capacity and a joint-space-dependent classification unit which is trained with cross-entropy loss to optimize classification performance. We evaluate models on full-resource and low- or zero-resource text classification of  multilingual news and biomedical text with a large label set. Our model  outperforms  monolingual and multilingual models which do not leverage label semantics and previous joint input-label space models in both scenarios.
\end{abstract}

\section{Introduction}

Text classification is a fundamental NLP task with numerous real-world applications such as topic recognition \cite{tang15,yang16}, sentiment analysis \cite{Pang2005,yang16}, and question answering \cite{Sukhbaatar15,kumar15}. Classification also appears as a sub task for sequence prediction tasks such as neural machine translation \cite{cho14,thang15},  and summarization \cite{rush15}. Despite the numerous studies, existing models are trained on a fixed label set using k-hot vectors and, therefore, treat target labels as mere atomic symbols without any particular structure to the space of labels,  ignoring potential linguistic knowledge about the words used to describe the output labels. Given that semantic representations of words have been shown to be useful for representing the input,  it is reasonable to expect that they are going to be useful for representing the labels as well.

Previous work has leveraged knowledge from the label texts through a joint input-label space, initially for image classification  \cite{Weston2011,mensik12,frome13,Socher13}.  Such models generalize to labels both seen and unseen during training, and scale well on very large label sets. However, as we explain in Section \ref{background}, existing input-label models for text \cite{yazdani15,Nam16} have the following limitations: (i) their embedding does not capture complex label relationships due to its bilinear form, (ii) their output layer parametrization is rigid because it depends on the dimensionality of the encoded text and labels, and, (iii) they are outperformed on seen labels by classification baselines trained with cross-entropy loss \cite{frome13,Socher13}. 

In this paper, we propose a new joint input-label model which generalizes over previous such models, addresses their limitations, and does not compromise performance on seen labels (see Figure \ref{schema-mini}). The proposed model is comprised of a joint non-linear input-label embedding with controllable capacity and a joint-space-dependent classification unit which is trained with cross-entropy loss to optimize classification performance.\footnote{Our code is available at:  \url{github.com/idiap/gile}} The need for capturing complex label relationships is addressed by two non-linear transformations which have the same target joint space dimensionality. The parametrization of the output layer is not constrained by the dimensionality of the input or label encoding, but is instead flexible with a capacity which can be easily controlled by choosing the dimensionality of the joint space.  Training is performed with cross-entropy loss, which is a suitable surrogate loss for classification problems, as opposed to a ranking loss such as WARP loss \cite{Weston10} which is more suitable for ranking problems.

Evaluation is performed on full-resource and low- or zero-resource scenarios of two text classification tasks, namely on biomedical semantic indexing \cite{Nam16} and on multilingual news classification \cite{pappas17c} against several competitive baselines. In both scenarios, we provide a comprehensive ablation analysis which highlights the importance of each model component and the difference with previous embedding formulations when using the same type of architecture and loss function.

Our main contributions are the following:

\begin{itemize}\addtolength{\itemsep}{-1ex}
\item[(i)] We identify key theoretical and practical limitations of existing joint  input-label models.

\item[(ii)] We propose a novel joint input-label embedding with flexible parametrization which generalizes over the previous such models and addresses their limitations.

\item[(iii)] We provide empirical evidence of the superiority of our model over monolingual and multilingual models which ignore label semantics, and over previous joint input-label models on both seen and unseen labels. 
\end{itemize}

The remainder of this paper is organized as follows. Section 2 provides background knowledge and explains limitations of existing models. Section 3 describes the model components, training and relation to previous formulations. Section 4 describes our evaluation results and analysis, while Section 5 provides an overview of previous work and  Section 6 concludes the paper and provides future research directions.

\section{Background: Neural Text Classification}
\label{background}
We are given a collection $D=\{(x_i, y_i), i=1, \ldots, N\}$ made of $N$ documents, where each document $x_i$ is associated with labels $y_i=\{y_{ij} \in \{0,1\}\ |\ j=1, \ldots, k\}$,  and k is the total number of labels.  Each document $x_{i}=\{w_{11},w_{12}, \ldots, w_{K_{i}T_{K_i}}\}$ is a sequence of words grouped into sentences, with $K_{i}$ being the number of sentences in document $i$ and $T_{j}$ being the number of words in sentence $j$.
Each label $j$ has a textual description comprised of multiple words, $c_{j} = \{c_{j1},c_{j2}, \dots, c_{jL_{j}}\ |\ j=1, \ldots, k \}$ with $L_{j}$ being the number of words in each description. 
Given the input texts and their associated labels \textit{seen} during the training portion of $D$, our goal is to learn a text classifier which is able to predict labels \textbf{both} in the seen, $\mathcal{Y}_s$, or unseen, $\mathcal{Y}_u$, label sets,  defined as the sets of unique labels which have been seen or not  during training respectively and, hence, $\mathcal{Y} \cap \mathcal{Y}_u = \emptyset$ and $\mathcal{Y} = \mathcal{Y}_s \cup \mathcal{Y}_u$.\footnote{Note that depending on the number of labels per document the problem can be a multi-label or multi-class problem.}

\subsection{Input Text Representation}
\label{input_enc}
To encode the input text, we focus on hierarchical attention networks (HANs), which are competitive for monolingual \cite{yang16} and multilingual text classification \cite{pappas17c}. The model takes as input a document $x$ and outputs a document vector $h$. The input words and label words are represented by vectors in $ \rm I\!R^d$ from the same\footnote{This statement holds true for multilingual classification problems too if the embeddings are aligned across languages.} embeddings $E \in \rm I\!R^{|\mathcal{V}|\times d}$, where $\mathcal{V}$ is the vocabulary and $d$ is the embedding dimension; $E$ can be pre-trained or learned jointly with the rest of the model. The model has two levels of abstraction, word and sentence. The word level is made of an encoder network $g_{w}$ and an attention network $a_{w}$, while the sentence level similarly includes an encoder and an attention network.  

\noindent\textbf{Encoders}. The function $g_w$ encodes the sequence of input words $\{w_{it}\ |\ t=1, \ldots, T_{i}\}$ for each sentence $i$ of the document, noted as:
\begin{gather}
  h^{(it)}_{w} = g_{w}(w_{it}), t \in [1, T_{i}],
\end{gather}
\noindent and at the sentence level, after combining the intermediate word vectors  $\{h^{(it)}_{w}\ |\  t=1, \ldots, T_{i}\}$ to a sentence vector $s_{i} \in \rm I\!R^{d_{w}}$ (see below), where $d_{w}$ is the  dimension of the word encoder, the function $g_s$ encodes the sequence of sentence vectors $\{s_{i}\ |\ i=1, \ldots, K\}$, noted as $h^{(i)}_{s}$. The $g_w$ and $g_s$ functions can be any feed-forward (\textsc{Dense}) or recurrent networks, 
e.g.~GRU \cite{cho14}.

\noindent\textbf{Attention}. The  $\alpha_{w}$ and $\alpha_{s}$ attention mechanisms, which estimate the importance of each hidden state vector, are used to obtain  the sentence $s_{i}$ and document representation $h$ respectively. The sentence vector is thus calculated as follows:
\begin{equation}
s_{i} = \sum^{T_{i}}_{t=1} \alpha^{(it)}_{w} h^{(it)}_{w} = \sum^{T_{i}}_{t=1} \frac{\mathrm{exp}(v^\top_{it} u_w)}{\sum_{j}\mathrm{exp}(v^\top_{ij} u_w)}  h^{(it)}_{w},
\label{att_eq}
\end{equation}
\noindent where $v_{it} = f_{w}(h^{(it)}_{w})$ is a fully-connected network with $W_w$ parameters. The document vector $h \in \rm I\!R^{d_{h}}$, where $d_{h}$ is the dimension of the sentence encoder, is calculated similarly, by replacing $u_{it}$ with $v_{i} = f_{s}(h^{(i)}_{s})$ which is a fully-connected network with $W_s$ parameters, and $u_w$ with $u_s$, which are parameters of the attention functions. 

\subsection{Label Text Representation}
To encode the label text we use an encoder function which takes as input a label description $c_j$ and outputs a label vector $e_j \in \rm I\!R^{d_{c}}\ \forall j=1, \ldots, k$.  For efficiency reasons, we use a simple, parameter-free function to compute $e_{j}$ , namely the average of word vectors which describe label $j$, namely $e_{j} = \frac{1}{L_{j}} \sum_{t=1}^{L_{j}} c_{jt}$, and hence $d_{c} = d$ in this case. By stacking all these label vectors into a matrix, we obtain the label embedding $\mathcal{E} \in \rm I\!R^{|\mathcal{Y}|\times d}$. In principle, we could also use the same encoder functions as the ones for input text, but this would increase the computation significantly; hence, we keep this direction as future work.

\subsection{Output Layer parametrizations}
\subsubsection{Typical Linear Unit}
The most typical output layer, consists of a linear unit with a weight matrix $W\in \rm I\!R^{d_h \times |\mathcal{Y}|}$ and a bias vector  $b  \in \rm I\!R^{|\mathcal{Y}|}$ followed by a softmax or sigmoid activation function. Given the encoder's hidden representation ${h}$ with dimension size $d_h$, the probability distribution of output $y$ given input $x$ is proportional to the following quantity:
\vspace{-2mm}
\begin{align}
 p(y|x) & \propto  \text{exp}(W^{\top}h + b).\label{baseline_full}
 \vspace{-2mm}
\end{align}
\noindent The parameters in $W$ can be learned separately or be tied with the parameters of the  embedding $E$ by setting $W = E^T$ if the input dimension of $W$ is restricted to be the same as that of the embedding $E$ ($d = d_h$) and each label is represented by a single word description i.e.~when $\mathcal{Y}$ corresponds to $\mathcal{V}$ and $E=\mathcal{E}$. In the latter case, Eq.~\ref{baseline_full} becomes:
\vspace{-2mm}
\begin{align}
  p(y|x) & \propto  \text{exp}(Eh + b). \label{baseline_eq}
	\vspace{-2mm}
\end{align}
Either way, the parameters of such models are typically learned with cross-entropy loss, which is suitable for classification problems.  However, in both cases they cannot be applied to labels which are not seen during training, because each label has learned parameters which are specific to that label, so the parameters for unseen labels cannot be learned.  We now turn our focus to a class of models which can handle unseen labels.

\subsubsection{Bilinear Input-Label Unit}
Joint input-output embedding models can generalize from seen to unseen labels because the parameters of the label encoder are shared.  The previously proposed joint input-output embedding models by \citet{yazdani15} and \citet{Nam16} are based on the following bilinear ranking function  $f(\cdot)$:
\begin{align}
  f(x, y) =  \mathcal{E} \mathcal{W}  h,
  \label{eq:bilinear}
\end{align}
where $\mathcal{E} \in \rm I\!R^{|\mathcal{Y}|\times d}$ is the label embedding and $\mathcal{W} \in \rm I\!R^{d \times d_h}$ is the bilinear embedding. This function allows one to define the rank of a given label $y$ with respect to $x$ and is trained using hinge loss to rank positive labels higher than negative ones.  But note that the use of this ranking loss means that they do not model the conditional probability, as do the traditional models above.

\noindent\textbf{Limitations}. Firstly, the above formula can only capture linear relationships between encoded text ($h$) and label embedding ($\mathcal{E}$) through $\mathcal{W}$. We argue that the relationships between different labels are non-linear due to the complex interactions of the semantic relations across labels but also between labels and different encoded inputs.  A more appropriate form for this purpose would include a non-linear transformation $\sigma(\cdot)$, e.g.~with either:
\begin{align}
  \text{(a)} \underbrace{\sigma(\mathcal{E} \mathcal{W})}_{\text{Label structure}}\hspace{-1ex}  h\ \
  \text{~~or~~}\ \ \text{(b)~~} \mathcal{E} \hspace{-1ex}\underbrace{\sigma(\mathcal{W} h)}_{\text{Input structure}}.
	\label{nonlin}
\end{align}
Secondly, it is hard to control their output layer capacity due to their bilinear form, which uses a  matrix of parameters ($\mathcal{W}$) whose size is bounded by the dimensionalities of the label embedding and the text encoding.
Thirdly, their loss function optimizes ranking instead of classification performance and thus treats the ground-truth as a ranked list when in reality it consists of one or more independent labels.

\noindent\textbf{Summary}. We hypothesize that these are the reasons why these models do not yet perform well on seen labels compared to models which make use of the typical linear unit, and they do not take full advantage of the structure of the problem when tested on unseen labels.  Ideally, we would like to have a model which will address these issues and will combine the benefits from both the typical linear unit and the joint input-label models.

 \section{The Proposed Output Layer Paramet- rization for   Text Classification}
\begin{figure}
 \centering
 \includegraphics[scale=0.4]{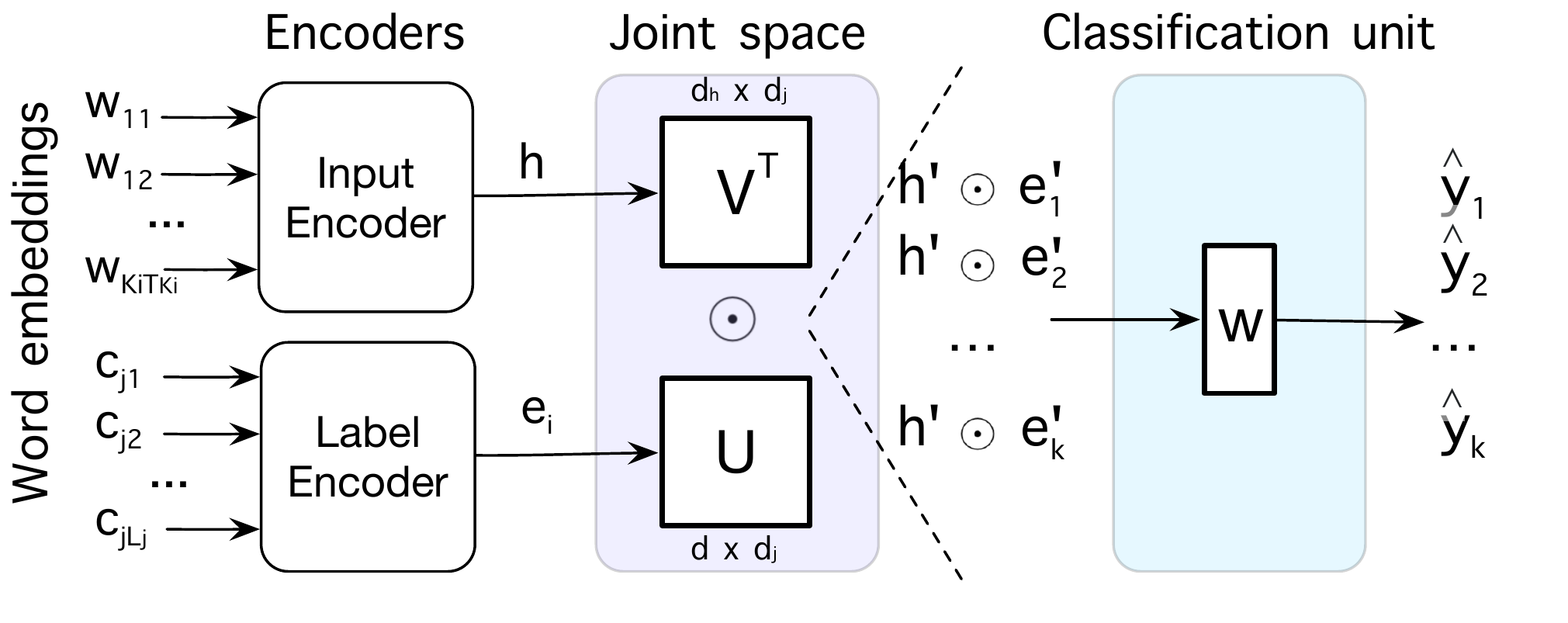}
 \caption{Each encoded text and label are projected to a joint input-label multiplicative space, the output of which is processed by a classification unit with label-set-size independent parametrization. }
 \label{schema-mini}
\end{figure}

 We propose a new output layer parametrization for neural text classification which is comprised of a generalized input-label embedding which captures the structure of the labels, the structure of the encoded texts and the interactions between the two, followed by  a classification unit which is independent of the label set size.  The resulting model has the following properties:  (i) it is able to capture complex output structure, (ii) it has a flexible parametrization which allows its capacity to be controlled, and (iii) it is trained with a classification surrogate loss such as cross-entropy. The model is depicted in Figure~\ref{schema-mini}. In this section, we describe the model in detail, showing how it can be trained efficiently for arbitrarily large label sets and how it is related to previous models.

\subsection{A Generalized Input-Label Embedding}
 Let $g_{in}(h)$ and $g_{out}(e_{j})$ be two non-linear projections of the encoded input, i.e.~the document $h$, and any encoded label $e_{j}$, where $e_{j}$ is the j$_{th}$ row vector from the label embedding matrix $\mathcal{E}$, which have the following form:
\begin{gather}
e_{j}' = g_{out}(e_{j}) = \sigma(e_{j} U  + b_{u})\\
h' = g_{in}(h) = \sigma( V h + b_{v}),
\end{gather}
\noindent where $\sigma(\cdot)$ is a nonlinear activation function such as ReLU or Tanh, the matrix $U \in \rm I\!R^{d \times d_j}$ and bias $b_{u} \in \rm I\!R^{d_j}$ are the linear projection of the labels, and the matrix $V \in \rm I\!R^{ d_j \times  d_h}$ and bias $b_{v} \in \rm I\!R^{d_j}$ are the linear projection of the encoded input.
Note that the projections for $h'$ and $e'_{j}$ could be high-rank or low-rank depending on their initial dimensions and the target joint space dimension.
Also let $\mathcal{E}' \in \rm I\!R^{|\mathcal{Y}| \times d_{j}} $ be the matrix resulting from projecting all the outputs $e_j$ to the joint space, i.e. $g_{out}(\mathcal{E})$.

The conditional output probability distribution can now be re-written as:
\begin{align}
 p(y|x) & \propto    \text{exp}\big( \mathcal{E}' h' \big)	 \nonumber\\
 & \propto  \text{exp}\big( g_{out}(\mathcal{E}) g_{in}(h) \big) \nonumber\\
 & \propto  \text{exp}\big( \underbrace{\sigma(\mathcal{E}U + b_{u})}_{\text{Label Structure}} \underbrace{\sigma( V h + b_{v})}_{\text{Input Structure}} \big).
 \label{general_form}
\end{align}
Crucially, this function has no label-set-size dependent parameters, unlike $\mathcal{W}$ and $b$ in Eq.~\ref{baseline_full}.
In principle, this parametrization can be used for both multi-class and multi-label problems by defining the exponential in terms of a softmax and sigmoid functions respectively. However, in this paper we will focus on the latter.

\subsection{Classification Unit} 
We require that our classification unit parameters depend only on the joint input-label space above. To represent the compatibility between any encoded input text $h_i$ and any encoded label $e_{j}$ for this task, we define their joint representation based on multiplicative interactions in the joint space:
\begin{equation}
g_{joint}^{(ij)} = g_{in}(h_i) \odot g_{out}(e_{j}),
\label{joint_space}
\end{equation}
where $\odot$ is component-wise multiplication.

The probability for $h_i$ to belong to one of the $k$ known labels is modeled by a linear unit which maps any point in the joint space into a score which indicates the validity of the combination:
\begin{equation}
  p_{val}^{(ij)} = g_{joint}^{(ij)} w + b,
	\label{prob}
\end{equation}
where $w \in \rm I\!R^{d_j}$ and $b$ are a scalar variables. We compute the output of this linear unit for each known label which we would like to predict for a given document $i$, namely:
\begin{equation}
  P_{val}^{(i)} =
  \begin{bmatrix}
    p^{(i1)}_{val}  \\
    p^{(i2)}_{val} \\
    \ldots \\
    p^{(ik)}_{val}
\end{bmatrix} =
\begin{bmatrix}
  g_{joint}^{(i1)} w + b  \\
  g_{joint}^{(i2)} w + b \\
  \ldots \\
  g_{joint}^{(ik)} w + b
\end{bmatrix}.
\end{equation}
For each row, the higher the value the more likely the label is to be assigned to the document. To obtain valid probability estimates and be able to train with binary cross-entropy loss for multi-label classification, we apply a sigmoid function as follows:
\begin{equation}
  \hat{y_i} = \hat{p}(y_i|x_i) = \frac{1}{1+e^{-P_{val}^{(i)} }}.
\end{equation}

\noindent \textbf{Summary}. By adding the above changes to the general form of Eq.~\ref{general_form} the conditional probability  $p(y_i|x_i)$ is now proportional to the following quantity:
\begin{equation}
    \text{exp}\big(  \sigma(\mathcal{E}U  + b_{u})   (\sigma( V h + b_{v})\odot w) + b   \big).
		\label{eq:out}
\end{equation}
Note that the number of parameters in this equation is independent of the size of the label set, given that $U$, $V$, $w$ and $b$ depend only on $d_{j}$, and $k$ can vary arbitrarily.  This allows the model to scale up to large label sets and generalize to unseen labels. Lastly, the proposed output layer addresses all the limitations of the previous models, as follows: (i) it is able to capture complex structure in the joint input-output space, (ii) it provides a means to easily control its capacity $d_{j}$, and (iii) it is trainable with cross-entropy loss.

\subsection{Training Objectives}

The training objective for the multi-label classification task is based on binary cross-entropy loss.  Assuming $\theta$ contains all the parameters of the model, the training loss is computed as follows:
\begin{equation}
\begin{split}
\mathcal{L}(\theta) & = - \frac{1}{Nk} \sum^N_{i=1} \sum_{j=1}^{k}  \mathcal{H}(y_{ij}, \hat{y}_{ij}), \label{eq:mono-objective} \\ 
\end{split}
\end{equation}
\noindent where $\mathcal{H}$ is the binary cross-entropy between the gold label $y_{ij}$ and predicted label $\hat{y}_{ij}$ for a document $i$ and a candidate label $j$.

We handle multiple languages according to \citet{firat16b} and  \citet{pappas17c}. Assuming that  $\Theta=\{\theta_1,\theta_2, ..., \theta_M\}$ are all the parameters required for each of the M languages, we use a joint multilingual objective based on the sum of cross-entropy losses:
\begin{equation}
  \label{multilingual_objective}
\begin{split}
  \mathcal{L}(\Theta)
 = - \frac{1}{Z} \sum^{N_e}_{i} \sum^{M}_{l} \sum_{j=1}^{k}  \mathcal{H}(y^{(l)}_{ij}, \hat{y}^{(l)}_{ij}),
\end{split}
\end{equation}
\noindent where $Z = N_e M  k$ with $N_e$ being the number of examples per epoch. At each iteration, a document-label pair for each language is sampled.  In addition, multilingual models share a certain subset of the encoder parameters during training while the output layer parameters are kept language-specific, as described by \citet{pappas17c}. In this paper, we share most of the output layer parameters, namely the ones from the input-label space (U, V, $b_v$, $b_u$), and we keep only the classification unit parameters ($w$, b) language-specific.

\subsection{Scaling Up to Large Label Sets}
For a very large number $d_j$ of joint-space dimensions in our parametrization, the computational complexity increases prohibitively because our projection requires a large matrix multiplication between $U$ and $E$, which depends on $|\mathcal{Y}|$. In such cases, we resort to sampling-based training, by adopting  the commonly used negative sampling method proposed by  \citet{mikolov13}. Let $x_i \in \rm I\!R^d$ and $y_{ik} \in \{0, 1\}$ be an input-label pair and $\hat{y_{ik}}$ the output probabilities from our model (Eq.~\ref{eq:out}). By introducing the sets $k_i^p$ and $k_i^n$, which contain the indices of the positive and negative labels respectively for the $i$-th input, the loss $L(\theta)$ in Eq.~\ref{eq:mono-objective} can be re-written as follows:
	\begin{align}  
   & = - \frac{1}{Z} \sum_{i=1}^N \sum_{j=1}^{k} \Big[ y_{ij} \log \hat{y}_{ij} + \bar{y}_{ij} \log\left(1 -\hat{y}_{ij}\right)\Big] \nonumber\\  
   \label{eq:loss_split}
	& =  -\frac{1}{Z} \sum_{i=1}^N \Big[~\sum_{j=1}^{k_i^p}\log\hat{y}_{ij} +
	    \sum_{j=1}^{k_i^n} \log\left(1 - \hat{y}_{ij} \right)\Big],
		\end{align}
\noindent where $Z = N k$ and $\bar{y}_{ij}$ is ($1 - y_{ij}$). To reduce the computational cost needed to evaluate $\hat{y}_{ij}$ for all the negative label set $k_i^n$, we sample $k^*$ labels from the negative label set with probability $p = \frac{1}{|k_i^n|}$ to create the set $k_i^n$. This enables training on arbitrarily big label sets without increasing the computation required. By controlling the number of samples we can drastically speed up the training time,
as we demonstrate  empirically in Section \ref{label-sampling}. Exploring more informative sampling methods, e.g.~importance sampling, would be an interesting direction of future work.

\subsection{Relation to Previous Parametrizations}
The proposed embedding form can be seen as a generalization over the input-label embeddings with a bilinear form, because its degenerate form is equivalent to the bilinear form of Eq.~\ref{eq:bilinear}. In particular, this can be simply derived if we set one of the two non-linear projection functions in the second line of Eq.~\ref{general_form} to be the identity function, e.g.~$g_{out}(\cdot) = I$, set all biases to zero, and make the $\sigma(.)$ activation function linear, as follows:
\begin{align}
  \sigma(\mathcal{E}U + b_{u})  \sigma( V h + b_{v})  & = ( \mathcal{E}I) \  (V h)  \nonumber\\
    & =   \mathcal{E}  V h\ \  \square,
\label{gen}
\end{align}
\noindent where $V$ by consequence has the same number of dimensions as $\mathcal{W} \in \rm I\!R^{d \times d_h}$ from the bilinear input-label embedding model of Eq.~\ref{eq:bilinear}.

\section{Experiments}
\label{sec:eval}
The evaluation is performed on large-scale biomedical semantic indexing using the BioASQ dataset, obtained by \citet{Nam16}, and on multilingual news classification using the DW corpus, which consists of eight language datasets obtained by \citet{pappas17c}. The statistics of these datasets are listed in Table \ref{tab_stats}.

\subsection{Biomedical Text Classification}
\label{bioexp}
We evaluate on biomedical text classification to demonstrate that our generalized  input-label model scales to very large label sets and performs better than previous joint input-label models on both seen and unseen label prediction scenarios.

\subsubsection{Settings}
We follow the exact evaluation protocol, data and settings of \citet{Nam16}, as described below. We use the BioASQ Task 3a dataset, which is a collection of scientific publications in biomedical research. The dataset contains about 12M documents labeled with around 11 labels out of 27,455, which are defined according to the Medical Subject Headings (MESH) hierarchy. 
The data was minimally pre-processed with tokenization, number replacements (NUM), rare word replacements (UNK), and split with the provided script by year so that the training set includes all documents until 2004 and the ones from 2005 to 2015 were kept for the test set, this corresponded to 6,692,815 documents for training and 4,912,719 for testing. For validation, a set of 100,000 documents were randomly sampled from the training set. We report the same ranking-based evaluation metrics as \citet{Nam16}, namely rank loss (RL), average precision (AvgPr) and one-error loss (OneErr).

\begin{table}[htp]
  \small
  \centering
    {\def\arraystretch{1.1}\tabcolsep=4pt
\begin{tabular}{ c | r r  r| r  r    }
  \hline
  \textbf{Dataset}  &   \multicolumn{3}{c|}{{   \textbf{Documents}}} &    \multicolumn{2}{c}{{   \textbf{Labels}}}   \\
     abbrev. & \# {count} & \# words & $\bar{w_d}$  & \# count & $\bar{w_l}$   \\ \hline\hline
  BioASQ & 11,705,534  & 528,156 & 214 & 26,104 & 35.0 \\ 
     DW & 598,304  & 884,272  &  436 & 5,637 &  2.3 \\
    -- en & 112,816 &  110,971 &  516 & 1,385 & 2.1\\
    -- de & 132,709 &  261,280 &  424 & 1,176 & 1.8  \\
    -- es & 75,827  &  130,661 &  412 & 843 & 4.7\\
    -- pt & 39,474  &  58,849 &  571 & 396 & 1.8\\
    -- uk & 35,423  &  105,240 &  342 & 288 & 1.7\\
    -- ru & 108,076 &  123,493 &  330 & 916 & 1.8 \\
    -- ar & 57,697  &  58,922 &  357 & 435 & 2.4\\
    -- fa & 36,282  &  34,856 &  538 & 198 & 2.5\\
\end{tabular}}
\caption{\label{tab_stats} Dataset statistics: \#count is the number of documents, \#words are the number of unique words in the vocabulary $\mathcal{V}$,  $\bar{w_d}$ and $\bar{w_l}$  are the average number of words per document and label respectively.}
\end{table}%

\begin{table*}
 \centering
 	{\def\arraystretch{1.2}\tabcolsep=2pt
\begin{tabular}{l lcc | ccc | ccc | r }\hline
 & \textbf{Model} & \textbf{Layer form} & \textbf{Dim} &  \multicolumn{3}{c |}{{   \textbf{Seen labels}}} &  \multicolumn{3}{c |}{{   \textbf{Unseen labels}}} & \textbf{Params} \\
 & abbrev. & output  & \#count & RL & AvgPr & OneErr & RL & AvgPr & OneErr & \#count  \\\hline \hline
  	\parbox[t]{3mm}{\multirow{3}{*}{\rotatebox[origin=c]{90}{ \textbf{[N16]}  }}} & WSABIE+  & $\mathcal{E} \mathcal{W}  h_{t} $ & 100 & 5.21 & 36.64 & 41.72 &  48.81 & 0.37 & 99.94 & 722.10M \\
  & AiTextML avg & $\mathcal{E} \mathcal{W}  h_{t} $ & 100 & 3.54 & 32.78 & 25.99 &   52.89 & 0.39 & 99.94 & 724.47M \\
	& AiTextML inf & $\mathcal{E} \mathcal{W}  h_{t} $ & 100 & 3.54 & 32.78 & 25.99 &  21.62 & 2.66 & 98.61 & 724.47M \\ \cdashline{1-11}
	\parbox[t]{3mm}{\multirow{3}{*}{\rotatebox[origin=c]{90}{ \textbf{Baselines}  }}} & WAN & $W^{\top}h_{t}$ & -- & 1.53  & 42.37 & \textbf{11.23} & -- & -- &  -- & 55.60M  \\
	& BIL-WAN [YH15]	& $\sigma(\mathcal{E}\mathcal{W}) \mathcal{W}  h_{t} $ & 100   & 1.21 & 40.68 & 17.52 & 18.72 & 9.50 & 93.89 & 52.85M  \\
  & BIL-WAN [N16]	& $\mathcal{E} \mathcal{W}  h_{t} $ & 100 & 1.12 & 41.91 & 16.94  & 16.26 & 10.55 & 93.23 & 52.84M  \\ \cline{1-11}
		\parbox[t]{3mm}{\multirow{4}{*}{\rotatebox[origin=c]{90}{ \textbf{Ours}  }}} & GILE-WAN   &$\sigma(\mathcal{E} U) \sigma( V h_{t} )$ &   500 & \textbf{0.78} & \textbf{44.39} & 11.60 & \textbf{9.06} & \textbf{12.95} & \textbf{91.90} & 52.93M  \\ \cline{2-11}
		& $-$ constrained $d_j$  & $\sigma(\mathcal{E} \mathcal{W}) \sigma( \mathcal{W} h_{t} )$ & 100 & 1.01 & 37.71 & 16.16 & 10.34 & 11.21 & 93.38 & 52.85M  \\
		& $-$ only label  (Eq.~\ref{nonlin}a) 	    & $\sigma(\mathcal{E} \mathcal{W})  h_{t} $ & 100 & 1.06   & 40.81 & 13.77  & 9.77 & 14.71 & 90.56  & 52.84M \\
  & $-$  only input (Eq.~\ref{nonlin}b) 	& $\mathcal{E} \sigma(\mathcal{W}  h_{t}) $ & 100 & 1.07  & 39.78 & 15.67 & 19.28 & 7.18 & 95.91 &   52.84M  \\
\end{tabular}}
\caption{Biomedical semantic indexing results computed over labels seen and unseen during training, i.e.\ the full-resource versus zero-resource settings. Best scores among the competing models are marked in \textbf{bold}.}
\label{biores}
\vspace{-3mm}
\end{table*}
Our hyper-parameters were selected on validation data based on average precision as follows: 100-dimensional word embeddings, encoder, attention (same dimensions as the  baselines),  joint input-label embedding of 500, batch size of 64, maximum number of 300 words per document and 50 words per label, ReLU activation,  0.3\% negative label sampling, and optimization with ADAM until convergence. The word embeddings were learned end-to-end on the task.\footnote{Here, the word embeddings are included in the parameter statistics because they are variables of the network.}

The baselines are the joint input-label models from \citet{Nam16}, noted as [N16], namely:
\begin{itemize}
	\item \textbf{WSABIE+}: This model is an extension of the original WSABIE model by \citet{Weston2011}, which instead of learning a ranking model with fixed document features, it jointly learns features for documents and words, and is trained with the WARP ranking loss.
\item  \textbf{AiTextML}: This model is the one proposed by \citet{Nam16} with the purpose of learning jointly representations of documents, labels and words, along with a joint input-label space which is trained with the WARP ranking loss.
\end{itemize}
The scores of the WSABIE+ and AiTextML baselines in Table \ref{biores} are the ones reported by \citet{Nam16}.
In addition, we report scores of a word-level attention neural network (WAN) with \textsc{Dense} encoder and attention followed by a sigmoid output layer, trained with binary cross-entropy loss.\footnote{ In our preliminary experiments, we also trained the neural model with a hinge loss as WSABIE+ and AiTextML, but it performed similarly to them and much worse than WAN, so we did not further experiment with it.  } Our model replaces WAN's output layer with a generalized input-label embedding layer and its variations, noted GILE-WAN.  For comparison, we also compare to bilinear input-label embedding versions of WAN for the model by \citet{yazdani15}, noted as BIL-WAN [YH16], and the one by \citet{Nam16}, noted as BIL-WAN [N16]. Note that the AiTextML parameter space is huge and makes learning difficult for our models (linear wrt. labels and documents). Instead, we make sure that our models have far fewer parameters than the baselines (Table \ref{biores}).

\subsubsection{Results}

The results on biomedical semantic indexing on seen and unseen labels are shown in Table \ref{biores}. We observe that the neural baseline, WAN, outperforms WSABIE+ and AiTextML on the seen labels, namely by $+$5.73 and $+$9.59 points in terms of AvgPr respectively. The differences are even more pronounced when considering the ranking loss and one error metrics.  This result is compatible with previous findings that existing joint input-label models are not able to outperform strong supervised baselines on seen labels. However, WAN is not able to generalize at all to unseen labels, hence the WSABIE+ and AiTextML have a clear advantage in the zero-resource setting.

In contrast, our generalized input-label model, GILE-WAN, outperforms WAN even on seen labels, where our model has higher average precision by $+$2.02 points, better ranking loss by $+$43\% and comparable OneErr ($-$3\%).  And this gain is not at the expense of performance on unseen labels.  GILE-WAN, outperforms WSABIE+, AiTextML variants\footnote{Namely, $avg$ when using the average of word vectors and \textit{inf} when using inferred label vectors to make predictions.}  by a large margin in both cases, e.g.~by  $+$7.75,  $+$11.61 points on seen labels and by  $+$12.58,  $+$10.29 points  in terms of average precision on unseen labels, respectively.  Interestingly, our GILE-WAN model also outperforms the two previous bilinear input-label embedding formulations of \citet{yazdani15}  and \citet{Nam16}, namely BIL-WAN [YH15] and BIL-WAN [N16] by $+$3.71,  $+$2.48  points on seen labels and  $+$3.45 and  $+$2.39 points on unseen labels, respectively, even when they are trained with the same encoders and loss as ours. These models are not able to outperform the WAN baseline when evaluated on the seen labels, namely they have $-$1.68 and $-$0.46 points lower average precision than WAN, but they outperform WSABIE+ and AiTextML on both seen and unseen labels. Overall, the results show a clear advantage of our generalized input-label embedding model against previous models on both seen and unseen labels.

\subsubsection{Ablation Analysis}
To evaluate the effectiveness of individual components of our model, we performed an ablation study (last three rows in Table \ref{biores}). Note that when we use only the label or only the input embedding in our generalized input-label formulation, the dimensionality of the joint space is constrained to be the dimensionality of the encoded labels and inputs respectively, that is~$d_j\text{=}100$ in our experiments.

All three variants of our model outperform previous embedding formulations of  \citet{Nam16} and  \citet{yazdani15} in all metrics except from AvgPr on seen labels where they score slightly lower. The decrease in AvgPrec for our model variants with $d_j\text{=}100$ compared to the neural baselines could be attributed to the difficulty in learning the parameters of a highly non-linear space with only a few hidden dimensions. Indeed, when we increase the number of dimensions ($d_j\text{=}500$), our full model outperforms them by a large margin. Recall that this increase in capacity is only possible with our full model definition in Eq.~9 and none of the other variants allow us to do this without interfering with the original dimensionality of the encoded labels ($\mathcal{E}$) and input ($h_t$). In addition, our model variants  with $d_j\text{=}100$ exhibit consistently higher scores than baselines in terms of most metrics on both seen and unseen labels, which suggests that they are able to capture more complex relationships across labels and between encoded inputs and labels.

Overall, the best performance among our model variants is achieved when using only the label embedding and, hence, it is the most significant component of our model. Surprisingly, our model with only the label embedding achieves higher performance than our full model on unseen labels but it is far behind our full model when we consider performance on both seen and unseen labels. When we constrain our full model to have the same dimensionality with the other variants, i.e.~$d_j\text{=}100$, it outperforms the one that uses only the input embedding in most metrics and it is outperformed by the one that uses only the label embedding.

\subsection{Multilingual News Text Classification}
We evaluate on multilingual news text classification to demonstrate that our output layer based on the generalized input-label embedding outperforms previous models with a typical output layer in a wide variety of settings, even for labels which have been seen during training.

\subsubsection{Settings}
\label{sec:exp-settings}
We follow the exact evaluation protocol, data and settings of \citet{pappas17c}, as described below. The dataset is split per language into 80\% for training, 10\% for validation and 10\% for testing. We evaluate on both types of labels (general $Y_{g}$, and specific $Y_{s}$) in a \textit{full-resource scenario},  and we evaluate only on the general labels ($Y_{g}$) in a \textit{low-resource scenario}. Accuracy is measured with the micro-averaged F1 percentage scores.
\begin{table*}[htp]
  \footnotesize
  \centering
  {\def\arraystretch{0.95}\tabcolsep=3.7pt
  \begin{tabular}{ p{0.2cm}  p{0.2cm} l | c c c c c c c | c c c c c c c | c}
  \hline
  \rowcolor{white}&    \multicolumn{2}{c|}{{   \textbf{Models}}}  & \multicolumn{7}{c|}{{   \textbf{Languages (en + aux $\rightarrow$ en)}}}  & \multicolumn{7}{c|}{{   \textbf{Languages (en + aux $ \rightarrow$ aux)} }} & \multicolumn{1}{c}{{   \textbf{Stat.} }} \\
   \rowcolor{white}  $Y_g$ &  \multicolumn{2}{c|}{{ abbrev. }} & de  & es  & pt & uk & ru & ar & fa & de & es & pt & uk & ru & ar & fa & \textit{avg} \\ \hline\hline
      \parbox[t]{1mm}{\multirow{6}{*}{\rotatebox[origin=c]{90}{  \textbf{[PB17]}  }}}
      &  \parbox[t]{1mm}{\multirow{3}{*}{\rotatebox[origin=c]{90}{ \textbf{Mono}  }}} & NN (Avg)  &  \multicolumn{7}{c|}{ 
50.7 \makebox[0.62\columnwidth]{\dotfill}
} & 53.1 & 70.0 & 57.2 & 80.9 & 59.3 & 64.4 & 66.6 & 57.6\\
      & & HNN (Avg) &  \multicolumn{7}{c|}{ 
70.0 \makebox[0.62\columnwidth]{\dotfill}
} & 67.9 & 82.5 & 70.5 & 86.8 & 77.4 & 79.0 & 76.6 & 73.6\\
      &  & HAN (Att)  & \multicolumn{7}{c|}{
71.2 \makebox[0.62\columnwidth]{\dotfill}
} & 71.8 & 82.8 & 71.3 & 85.3 & 79.8 & 80.5 & 76.6 & 74.7 \\
       & \parbox[t]{2mm}{\multirow{3}{*}{\rotatebox[origin=c]{90}{  \textbf{Multi}  }}} &  MHAN-Enc  & 71.0 & 69.9 & 69.2 & \textbf{{70.8}} & 71.5 & 70.0 & 71.3 & 69.7 & {82.9} & 69.7 & 86.8 & 80.3 & 79.0 & 76.0 & 74.1\\
    & &  MHAN-Att &  {74.0} & {74.2} & {74.1} & {72.9} & {73.9} & {73.8} & {73.3} & {72.5} & 82.5 & 70.8 & {87.7} & 80.5 & {82.1} & 76.3  & \underline{76.3}\\
    & &  MHAN-Both &  72.8 & 71.2 & 70.5 & 65.6 & 71.1 & 68.9 & 69.2 & 70.4 & 82.8 & {71.6} & 87.5 & {80.8} & 79.1 & {77.1} & 74.2 \\ \hline

    \parbox[t]{1mm}{\multirow{6}{*}{\rotatebox[origin=c]{90}{  \textbf{Ours} }}} &  \parbox[t]{2mm}{\multirow{3}{*}{\rotatebox[origin=c]{90}{  \textbf{Mono}  }}}  &  GILE-NN (Avg)  & \multicolumn{7}{c|}{
\textbf{60.1} \makebox[0.62\columnwidth]{\dotfill}
}  & \textbf{60.3} & \textbf{76.6}  & \textbf{62.1} & \textbf{82.0} & \textbf{65.7}  & \textbf{77.4}   & \textbf{68.6} & \textbf{65.2}\\
     & & GILE-HNN (Avg) &  \multicolumn{7}{c|}{
  \textbf{74.8} \makebox[0.62\columnwidth]{\dotfill}
} &  \textbf{71.3} & \textbf{83.3} & \textbf{72.6} &  \textbf{88.3}  & \textbf{81.5} &  \textbf{81.9}   &  \textbf{77.1} & \textbf{77.1} \\
      &  & GILE-HAN (Att)  &  \multicolumn{7}{c|}{
{\textbf{76.5}} \makebox[0.62\columnwidth]{\dotfill}
}  & {\textbf{74.2}} & {\textbf{83.4}}  & \textbf{71.9} & \textbf{86.1}  & \textbf{82.7}  & \textbf{81.0}  &  \textbf{77.2}  & \textbf{78.0} \\
    &  \parbox[t]{2mm}{\multirow{3}{*}{\rotatebox[origin=c]{90}{  \textbf{Multi}  }}}  & GILE-MHAN-Enc  & \textbf{75.1} & \textbf{74.0} &\textbf{72.7} &  70.7 & \textbf{74.4}&  \textbf{73.5}  & \textbf{73.2}  & \textbf{72.7} & \textbf{83.4} & {\textbf{73.0}} & {\textbf{88.7}} & \textbf{82.8} & {\textbf{83.3}}  & {\textbf{77.4}} & \textbf{76.7} \\
  & & GILE-MHAN-Att & \textcolor{black}{\textbf{76.5}} & \textcolor{black}{\textbf{76.5}}& \textcolor{black}{\textbf{76.3}}& \textcolor{black}{\textbf{75.3}} & \textcolor{black}{\textbf{76.1}}& \textcolor{black}{\textbf{75.6}}& \textcolor{black}{\textbf{75.2}} & \textcolor{black}{\textbf{74.5}} & {\textbf{83.5}} & \textcolor{black}{\textbf{72.7}} &  \textbf{88.0} & \textcolor{black}{\textbf{83.4}}& \textcolor{black}{\textbf{82.1}} & {\textbf{76.7}}  & \underline{\textbf{78.0}}\\
  & &  GILE-MHAN-Both & \textbf{75.3}  &  \textbf{73.7} & \textbf{72.1} & \textbf{67.2} & \textbf{72.5} &\textbf{73.8}  & \textbf{69.7} & \textbf{72.6} & \textcolor{black}{\textbf{84.0}}  & \textcolor{black}{\textbf{73.5}} & \textcolor{black}{\textbf{89.0}} & \textbf{81.9} & \textbf{82.0} & \textcolor{black}{\textbf{77.7}} & {\textbf{76.0}} \\
  \hline   \hline
  \rowcolor{white}   $Y_\mathit{s}$ & & Models & de  & es  & pt & uk & ru & ar & fa & de & es & pt & uk & ru & ar & fa & \textit{avg} \\ \hline \hline
  \parbox[t]{1mm}{\multirow{6}{*}{\rotatebox[origin=c]{90}{ \textbf{[PB17]} }}} &  \parbox[t]{2mm}{\multirow{3}{*}{\rotatebox[origin=c]{90}{  \textbf{Mono}  }}}  &  NN (Avg)  & \multicolumn{7}{c|}{
24.4 \makebox[0.62\columnwidth]{\dotfill}
}  & 21.8 & 22.1 & 24.3 & 33.0 & 26.0 & 24.1 & 32.1 & 25.3 \\
   & & HNN (Avg) &  \multicolumn{7}{c|}{
39.3 \makebox[0.62\columnwidth]{\dotfill}
} & 39.6 & 37.9 & 33.6 & 42.2 & 39.3 & 34.6 & 43.1 & 38.9\\
    &  & HAN (Att)  &  \multicolumn{7}{c|}{
43.4 \makebox[0.62\columnwidth]{\dotfill}
} & 44.8 & 46.3 & 41.9 & 46.4 & 45.8 & 41.2 & {49.4} & 44.2 \\
  &  \parbox[t]{2mm}{\multirow{3}{*}{\rotatebox[origin=c]{90}{  \textbf{Multi}   }}}  & MHAN-Enc  & 45.4 & 45.9 & {44.3} & 41.1 & 42.1 & \textbf{44.9} & 41.0 & 43.9 & 46.2 & 39.3 & 47.4 & 45.0 & 37.9 & 48.6  & 43.8\\
& & MHAN-Att &  {46.3} & {46.0} & \textbf{45.9} & {45.6} & {46.4} & {46.4} & \textbf{46.1} & {46.5} & {46.7} & {43.3} & {47.9} & {45.8} & {41.3} & 48.0 & \underline{45.8} \\
& &  MHAN-Both &  45.7 & 45.6 & 41.5 & 41.2 & 45.6 & \textbf{44.6} & \textbf{43.0} & 45.9 & 46.4 & 40.3 & 46.3 & \textbf{46.1} & 40.7 & \textbf{50.3} & 44.5\\
\hline
\parbox[t]{1mm}{\multirow{6}{*}{\rotatebox[origin=c]{90}{ \textbf{Ours} }}} &  \parbox[t]{2mm}{\multirow{3}{*}{\rotatebox[origin=c]{90}{  \textbf{Mono}  }}}  & GILE-NN (Avg)  & \multicolumn{7}{c|}{
\textbf{27.5} \makebox[0.62\columnwidth]{\dotfill}
}  & \textbf{27.5} & \textbf{28.4}  & \textbf{29.2}  & \textbf{36.8}  & \textbf{31.6}  & \textbf{32.1}  & \textbf{35.6} &  \textbf{29.5}\\
 & & GILE-HNN (Avg) &  \multicolumn{7}{c|}{
 \textbf{43.1} \makebox[0.62\columnwidth]{\dotfill}
} &  \textbf{43.4} & \textbf{42.0}  &  \textbf{37.7}  &   \textbf{43.0}  & \textbf{42.9} &  \textbf{36.6}  &  \textbf{44.1} & \textbf{42.2}\\
  &  & GILE-HAN (Att)  &  \multicolumn{7}{c|}{
 \textbf{45.9}  \makebox[0.62\columnwidth]{\dotfill}
}  & \textbf{47.3} &  \textbf{47.4} & \textbf{42.6} &  \textbf{46.6}   & \textbf{46.9}  & \textbf{41.9}  & 48.6 & \textbf{45.9}\\
&  \parbox[t]{2mm}{\multirow{3}{*}{\rotatebox[origin=c]{90}{  \textbf{Multi}  }}}  & GILE-MHAN-Enc  & \textbf{46.0} &  \textbf{46.6} & 41.2  & \textbf{42.5} & {\textbf{46.4}}  & 43.4 & \textbf{41.8} & \textbf{47.2} &  \textbf{47.7} & \textbf{41.5}  & {\textbf{49.5}} & {\textbf{46.6}} &  \textbf{41.4} & {\textbf{50.7}} & \textbf{45.1}\\
& & GILE-MHAN-Att & {\textbf{47.3}} & {\textbf{47.0}} & {\textbf{45.8}} & {\textbf{45.5}} & \textbf{46.2} & {\textbf{46.5}} & 45.5 & {\textbf{47.6}} & {\textbf{47.9}} & {\textbf{43.5}}& \textbf{49.1} & \textbf{46.5} &{\textbf{42.2}}  & \textbf{50.3} & \underline{\textbf{46.5}}\\
& &  GILE-MHAN-Both& \textbf{47.0} &  {\textbf{46.7}} & {\textbf{42.8}} & {\textbf{42.0}} & \textbf{45.6}  & 42.8  & 39.3 & {\textbf{48.0}} & \textbf{47.6} &\textbf{43.1}  & \textbf{48.5} & 46.0 & \textbf{42.1} & 49.0 & \textbf{45.0} \\
\hline
\end{tabular}}
\caption{\label{full-rec_results}Full-resource classification results on general (upper half) and specific (lower half) labels using monolingual and bilingual models with \textsc{Dense} encoders on English as target (left) and the auxiliary language as target (right).  The average bilingual F1-score (\%) is noted  \textit{avg} and the top ones per block are \underline{underlined}. The monolingual scores on the left come from a single model, hence a single score is repeated multiple times; the repetition is marked with consecutive dots.
}
  \vspace{-3mm}
\end{table*}

The word embeddings for this task are the aligned pre-trained 40-dimensional multi-CCA multilingual word embeddings by \citet{ammar16} and are kept fixed during training.\footnote{The word embeddings are not included in the parameters statistics   because they are not variables of the network.} The sentences are already truncated at a length of 30 words and the documents at a length of 30 sentences. The hyper-parameters were selected on validation data as follows: 100-dimensional encoder and attention, $\mathrm{ReLU}$ activation, batch size of 16, epoch size of 25k, no negative sampling (all labels are used) and optimization with ADAM until convergence. To ensure equal capacity to baselines, we use approximately the same number of parameters $n_{tot}$ with the baseline classification layers, by setting:
\begin{equation}
  d_{j} \simeq   \frac{d_h * |k^{(i)}|}{{d_h + d}}, \ i=1,\ldots, M,
\end{equation}
in the monolingual case, and similarly, $d_{j} \simeq   (d_h * \sum_{i=1}^M |k^{(i)}|) / ({d_h + d})$  in the multilingual case, where $k^{(i)}$ is the number of labels in language $i$.

The hierarchical models have \textit{Dense} encoders in all scenarios (Tables~\ref{full-rec_results},~\ref{multi8},  and~\ref{low-rec_results}), except from the varying encoder experiment (Table~\ref{encoders}). For the low-resource scenario, the levels of  data availability are:  \textit{tiny} from 0.1\% to 0.5\%, \textit{small} from 1\% to 5\% and \textit{medium} from 10\% to 50\% of the original training set. For each level, the average F1 across discrete increments of 0.1, 1 and 10 are reported respectively. The decision thresholds, which were tuned on validation data by \citet{pappas17c}, are set as follows: for the full-resource scenario it is set to 0.4 for $|Y_{s}|<400$ and 0.2 for $|Y_{s}|\geq400$, and for the low-resource scenario it is set to 0.3 for all sets.

The baselines are all the monolingual and multilingual neural networks from \citet{pappas17c}\footnote{For reference, in Table \ref{encoders} we also compare to a logistic regression  trained with unigrams over the full vocabulary and over the top-10\% most frequent words by \citet{mrini17}, noted as [M17], which use the same settings and data.}, noted as [PB17], namely:
\begin{itemize} \setlength{\itemsep}{-0.05em}
 \item \textbf{NN}: A neural network which feeds the average vector of the input words directly to a classification layer, as the one used by \citet{klementiev12}.
 \item \textbf{HNN}: A hierarchical network with encoders and average pooling at every level,	followed by a classification layer, as the one used by \citet{tang15}.
 \item  \textbf{HAN}: A hierarchical network with encoders and attention, followed by a classification layer, as the one used by \citet{yang16}.
 \item  \textbf{MHAN}: Three multilingual hierarchical networks with shared encoders, noted MHAN-Enc, shared attention, noted MHAN-Att, and shared attention and encoders, noted MHAN-Both, as the ones used by \citet{pappas17c}.
 \end{itemize}

To ensure a controlled comparison to the above baselines, for each model we evaluate a version where their output layer is replaced by our generalized input-label embedding output layer using the same number of parameters; these  have the abbreviation ``GILE'' prepended in their name (e.g.~GILE-HAN). The scores of HAN and MHAN models in Tables \ref{full-rec_results}, \ref{multi8} and \ref{low-rec_results}  are the ones reported by \citet{pappas17c}, while for Table \ref{encoders} we train them ourselves using their code. Lastly, the best score for each pairwise comparison between a joint input-label model and its counterpart is marked in \textbf{bold}.

\begin{table*}
  \def\arraystretch{1.2}
  \setlength{\tabcolsep}{4pt}
  \centering
  \small
\begin{tabular}{ p{4mm}  l |   c  c  c  c  c  c c  c | c  c } \hline
  &    \multicolumn{1}{c|}{{   \textbf{Models}}}  & \multicolumn{8}{c|}{{   \textbf{Languages}}}   & \multicolumn{2}{c}{{   \textbf{Statistics}}}  \\
        &  \multicolumn{1}{c|}{{ abbrev. }} &   en &  de &  es & pt & uk & ru & ar & fa & $n_l$ & $f_l$  \\ \hline
      \hline \parbox[t]{1mm}{\multirow{2}{*}{\rotatebox[origin=c]{90}{  \textbf{[M17] }  }}}  & LogReg-\textsc{bow}  & 75.8 & 72.9  & 81.4  & 74.3  & \textcolor{black}{91.0}  & 79.2 & 82.0 & 77.0 & 26M  & 79.19  \\
				& LogReg-\textsc{bow}-\textsc{10\%} & 74.7 & 70.1  & 80.6  & 71.1  & \textcolor{black}{89.5}  & 76.5 & 80.8 & 75.5 & 5M  & 77.35 \\ \cdashline{1-12}
     \parbox[t]{1mm}{\multirow{3}{*}{\rotatebox[origin=c]{90}{  \textbf{[PB17] }  }}}    & HAN-\textsc{bigru}     &  76.3  & \textbf{74.1} & 84.5 & \textcolor{black}{\textbf{72.9}}  & 87.7 & \textcolor{black}{\textbf{82.9}} & 81.7  & {75.3}   &  377K \textcolor{black}{}  & 79.42 \textcolor{black}{}   \\
 & HAN-\textsc{gru}    &  \textbf{77.1}  & 72.5 &  84.0 & 70.8 & 86.6 & 83.0 & 82.9  & 76.0  & 138K   &  79.11 \textcolor{black}{} \\
 & HAN-\textsc{Dense}     & 71.2  & 71.8 & 82.8 & 71.3 & 85.3 & 79.8 & 80.5 & 76.6  & 50K   & 77.41 \textcolor{black}{} \\
     \hline
      \parbox[t]{1mm}{\multirow{3}{*}{\rotatebox[origin=c]{90}{  \textbf{Ours}  }}}
      & GILE-HAN-\textsc{bigru}  & \textcolor{black}{\textbf{78.1}} & 73.6 & \textcolor{black}{\textbf{84.9}} & 72.5  &\textcolor{black}{\textbf{89.0}} & 82.4 & \textcolor{black}{\textbf{82.5}}   & \textcolor{black}{\textbf{75.8}}    &  377K \textcolor{black}{} & \textbf{79.85}   \\
      & GILE-HAN-\textsc{gru} & {77.1} & \textbf{72.6} & \textbf{84.7} & \textbf{72.4} & \textbf{88.6} & \textbf{83.6} & \textbf{83.4} &  \textbf{76.0} & 138K  & \textbf{79.80}    \\
       & GILE-HAN-\textsc{Dense}  & \textbf{76.5} & \textcolor{black}{\textbf{74.2}} & \textbf{83.4} & \textbf{71.9} & \textbf{86.1} & \textbf{82.7} & \textbf{82.6} & \textbf{77.2} & 50K  & \textbf{79.12}     \\ \hline
 \end{tabular}
\caption{\label{encoders} Full-resource classification results on general ($Y_g$) topic labels with \textsc{Dense} and \textsc{gru} encoders. Reported are also the average number of parameters per language ($n_{l}$), and the average $F_1$ per  language ($f_{l}$). 
}
\end{table*}

\subsubsection{Results}
\label{dwres}
Table~\ref{full-rec_results} displays the results of full-resource document classification using \textsc{Dense} encoders for both general and specific labels.  On the left, we display the performance of models on the English sub-corpus  when  English  and  an auxiliary language are used for training, and on the right,  the  performance  on  the  auxiliary language sub-corpus when that language and English are used for training.

The results show that in 98\% of comparisons on general labels  (top half of Table~\ref{full-rec_results}) the joint input-label models improve consistently over the corresponding models using a typical sigmoid classification layer. This finding validates our main hypothesis that the joint input-label models successfully  exploit the semantics of the labels, which provide useful cues for classification, as opposed to models which are agnostic to label semantics. The results for specific labels (bottom half of Table~\ref{full-rec_results}) demonstrate the same trend, with the joint input-label models performing better in 87\% of comparisons. 

In Table \ref{direct}, we also directly compare our embedding to previous bilinear input-label embedding formulations when using the best monolingual configuration (HAN) from Table \ref{full-rec_results}, exactly as done in Section \ref{bioexp}. The results on the general labels show that GILE outperforms the previous bilinear input-label models, BIL [YH15] and BIL [N16], by +1.62 and +3.3 percentage points on average respectively. This difference is much more pronounced on the specific labels, where the label set is much larger, namely +6.5 and +13.5 percentage points respectively. Similarly, our model with constrained dimensionality is also as good or better on average than the bilinear input-label models, by +0.9 and +2.2 on general labels and by -0.5 and +6.1 on specific labels respectively, which highlights the importance of learning non-linear relationships across encoded labels and documents. Among our ablated model variants, as in previous section, the best is the one with only the label projection but it still worse than our full model by -5.2 percentage points.  The improvements of GILE against each baseline is significant and consistent on both datasets. Hence, in the following experiments we will only consider the best of these alternatives.
\begin{table}[ht]
 \vspace{-1mm}
  \def\arraystretch{1.4}
  \setlength{\tabcolsep}{2pt}
  \centering
  \footnotesize
\begin{tabular}{ l | c  c  c  c  c  c c  c  } \hline
    \multicolumn{1}{c|}{{   \textbf{HAN }}}  & \multicolumn{8}{c}{{   \textbf{Languages}}}    \\
       \multicolumn{1}{l|}{{$Y_g$ output layer }} &   en &  de &  es & pt & uk & ru & ar & fa \\ \hline \hline
  Linear [PB17]  & 71.2  & 71.8 & 82.8 & 71.3 & 85.3 & 79.8 & 80.5 & 76.6  \\ 
         BIL [YH15]   & 71.7  & 70.5  & 82.0 & 71.1  & 86.6  & 80.6 & 80.4 & 76.0  \\ 
         BIL [N16]   & 69.8  & 69.1  & 80.9 & 67.4  & \textbf{87.5}  & 79.9 & 78.4 & 75.1  \\ \hline 
                 GILE (Ours)  & \textbf{76.5} & \textcolor{black}{\textbf{74.2}} & \textbf{83.4} & \textbf{71.9} & {86.1} & \textbf{82.7} & \textbf{82.6} & \textbf{77.2}  \\ \hline
                - constrained$\,d_j\!\!$  & 73.6  & 73.1  & 83.3 & 71.0  & 87.1  & 81.6  & 80.4 & 76.4  \\ 
                - only label  & 71.4  & 69.6  & 82.1 & 70.3  & {86.2}  & 80.6  & 81.1 & 76.2   \\
                - only input   & 55.1  & 54.2  & 80.6 & 66.5  & 85.6  & 60.8 & 78.9 &  74.0 \\
    \hline \hline \\[-3ex]
   \multicolumn{1}{l|}{{$Y_s$ output layer  }} &   en &  de &  es & pt & uk & ru & ar & fa \\ \hline \hline
    Linear[PB17]     & 43.4  & 44.8 & 46.3 & 41.9 & 46.4 & 45.8 & 41.2 & \textbf{49.4}  \\
        BIL [YH15]  & 40.7 &  37.8 & 38.1 & 33.5 &  44.6 & 38.1 & 39.1 & 42.6  \\
         BIL [N16]  & 34.4 & 30.2 & 34.4 & 33.6 & 31.4 & 22.8 & 35.6  & 38.9  \\ \hline
                 GILE  (Ours)  & \textbf{45.9} & \textcolor{black}{\textbf{47.3}} & \textbf{47.4} & \textbf{42.6} & \textbf{46.6} & \textbf{46.9} & \textbf{41.9} & {48.6}  \\ 
                 \hline
         - constrained$\,d_j\!\!$  & 38.5  &  38.0 & 36.8 & 35.1   & 42.1 & 36.1 & 36.7 & 48.7  \\
         - only label  & 38.4  & 41.5  & 42.9 & 38.3  & 44.0  & 39.3 & 37.2 & 43.4  \\
         - only input  & 12.1  & 10.8  & 8.8 & 20.5 & 11.8  & 7.8 & 12.0 & 24.6  \\
 \end{tabular}
\caption{\label{direct} Direct comparison with previous bilinear input-label models, namely BIL [YH15] and BIL [N16], 
and with our ablated model variants using the best monolingual configuration (HAN) from Table \ref{full-rec_results} on both general (upper half) and specific (lower half) labels. Best scores among the competing models are marked in \textbf{bold}.
}
 \vspace{-4mm}
\end{table}

The best bilingual performance on average is that of the GILE-MHAN-Att model, for both general and specific labels. This improvement can be attributed to the effective sharing between label semantics across languages through the joint multilingual input-label output layer. Effectively, this model has the same multilingual sharing scheme  with the best model reported by \citet{pappas17c}, MHAN-Att, namely sharing attention at each level of the hierarchy, which agrees well with their main finding.
\begin{table}[htp]
  \centering
  \small
  {\def\arraystretch{1.15}\tabcolsep=1pt
\begin{tabular}{ p{3mm}  l   c |  c  c  |  c  c }\hline
  &   \multicolumn{2}{c|}{\textbf{Models} }  & \multicolumn{2}{c|}{{   \textbf{General labels}}}   & \multicolumn{2}{c}{ \textbf{Specific labels} }  \\
      \rowcolor{white}   &   \multicolumn{1}{c}{{ abbrev. }}   & \# lang.  &   $ n_{l}$ & $f_{l}$  &   $ n_{l}$ & $f_{l}$  \\ \hline     \hline \parbox[t]{1mm}{\multirow{3}{*}{\rotatebox[origin=c]{90}{  \textbf{[PB17]}  }}}  & HAN   &  1 & 50K   & 77.41  & 90K   & 44.90 \\
         & MHAN &  2 & 40K  & 78.30  & 80K  & 45.72  \\
      &  MHAN   &  8 & 32K  & 77.91   & 72K & 45.82  \\ \hline
      \parbox[t]{1mm}{\multirow{3}{*}{\rotatebox[origin=c]{90}{  \textbf{Ours}  }}}  & GILE-HAN  &  1 &  50K  & \textbf{79.12}   &   90K &   \textbf{45.90}   \\
         & GILE-MHAN &  2 &  40K  & \textbf{79.68}   &  80K &  \textcolor{black}{\textbf{46.49}}  \\
      & GILE-MHAN &  8 & 32K & \textcolor{black}{\textbf{79.48}}   &  72K & \textbf{46.32}   \\\hline
 \end{tabular}}
\caption{\label{multi8} Multilingual learning results. The columns are the average number of parameters per language ($n_{l}$), average $F_1$ per  language ($f_{l}$). 
}
\end{table} 
Interestingly, the improvement holds when using different types of hierarchical encoders, namely \textsc{Dense} GRU, and biGRU, as shown in Table \ref{encoders}, which demonstrate the generality of the approach. In addition, our best models outperform logistic regression trained either on top-10\% most frequent words or on the full vocabulary, even though our models utilize many fewer parameters, namely 377K/138K vs.~26M/5M. Increasing the capacity of our models should lead to even further improvements.

\textbf{Multilingual learning.}
So far, we have shown that the proposed joint input-label models outperform typical neural models when training with one and two languages.
Does the improvement remain when increasing the number of languages even more? To answer the question we report in Table \ref{multi8} the average F1-score per language for the best baselines from the previous experiment (HAN and MHAN-Att) with the proposed joint input-label versions of them (GILE-HAN and GILE-MHAN-Att) when increasing the number of languages (1, 2 and 8) that are used for training. Overall, we  observe that the joint input-label models outperform all the baselines independently of the number of languages involved in the training, while having the same number of parameters.
 We also replicate the previous result that a second language helps but beyond that there is no improvement.

\textbf{Low-resource transfer.}
We investigate here whether joint input-label models are useful for low-resource languages. Table \ref{low-rec_results} shows the low-resource classification results from English to seven other languages when varying the amount of their training data. Our model with both shared encoders and attention, GILE-MHAN, outperforms previous models in average, namely HAN \cite{yang16} and MHAN \cite{pappas17c}, for low-resource classification in the majority of the cases. %

\begin{table}[ht]
  \small
\centering
  {\def\arraystretch{1.1}\tabcolsep=6pt
  \begin{tabular}{ c | c | c   c | c  }
    \hline
    & \textbf{Levels}  & \multicolumn{2}{c|}{\textbf{[PB17]}}   & \textbf{Ours} \\
    \rowcolor{white}    & range &  HAN & MHAN & GILE-MHAN \\
    \hline\hline
     \parbox[t]{3mm}{\multirow{3}{*}{\rotatebox[origin=c]{90}{{en}$\rightarrow${de}}}} & 0.1-0.5\% & 29.9 & 39.4 & \textbf{42.9}\\
     & 1-5\% & 51.3 & \textbf{52.6}  & 51.6 \\
     & 10-50\% & {63.5} & {63.8}  & \textbf{65.9}\\ \hline

     \parbox[t]{2mm}{\multirow{3}{*}{\rotatebox[origin=c]{90}{{en}$\rightarrow${es}}}} & 0.1-0.5\% & 39.5  & \textbf{41.5} & 39.0 \\
      & 1-5\%     & 45.6  & 50.1 & \textbf{50.9} \\
      & 10-50\%   & 74.2  & 75.2 & \textbf{76.4} \\ \hline

     \parbox[t]{2mm}{\multirow{3}{*}{\rotatebox[origin=c]{90}{{en}$\rightarrow${pt}}}} & 0.1-0.5\% & 30.9 &   {33.8} & \textbf{39.6} \\
     & 1-5\% & 44.6 &   {47.3} & \textbf{48.9}   \\
     & 10-50\% & 60.9 &   {62.1} & \textbf{62.3} \\ \hline

     \parbox[t]{3mm}{\multirow{3}{*}{\rotatebox[origin=c]{90}{{en}$\rightarrow${uk}}}} & 0.1-0.5\% & 60.4 & 60.9 & \textbf{61.1}\\
      & 1-5\%     & 68.2 &  69.0 & \textbf{69.4}\\
      & 10-50\%   & 76.4 & \textbf{76.7} & 76.5\\ \hline

     \parbox[t]{2mm}{\multirow{3}{*}{\rotatebox[origin=c]{90}{{en}$\rightarrow${ru}}}} & 0.1-0.5\% & 27.6 &  \textbf{29.1} & 27.9 \\
      & 1-5\%     & 39.3 & \textbf{40.2} & {40.2}\\
      & 10-50\%   & 69.2  & {69.4} & \textbf{70.4}\\ \hline

     \parbox[t]{2mm}{\multirow{3}{*}{\rotatebox[origin=c]{90}{{en}$\rightarrow${ar}}}} & 0.1-0.5\% & 35.4 & 36.6 & \textbf{46.1}\\
     & 1-5\%      & 45.6 & 46.6 & \textbf{49.5}\\
     & 10-50\%    & 48.9 & 47.8 & \textbf{61.8} \\ \hline

     \parbox[t]{3mm}{\multirow{3}{*}{\rotatebox[origin=c]{90}{{en}$\rightarrow${fa}}}} & 0.1-0.5\% & 36.0 & {41.3} & \textbf{42.5}\\
     & 1-5\%      & 55.0 & \textbf{55.5} & 55.4 \\
     & 10-50\%    & 69.2 & \textbf{70.0} &  69.7\\ \hline
  \end{tabular}}
\caption{\label{low-rec_results}Low-resource classification results with various sizes of training data using the general labels.
}
\end{table}
The shared input-label space appears to be helpful especially when transferring from English to German, Portuguese and Arabic languages. GILE-MHAN is significantly behind MHAN on transferring knowledge from English to Spanish and to Russian in the 0.1-0.5\% resource setting, but in the rest of the cases they have very similar scores.

\begin{figure}[htp]
\centering
\includegraphics[scale=0.38]{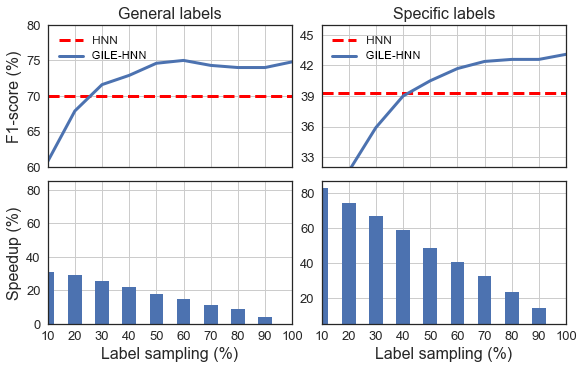}
\caption{Varying sampling percentage for general and specific English labels.
\textit{(Top)} GILE-HNN is compared against HNN in terms of F1 (\%). \textit{(Bottom)} The runtime speedup over GILE-HNN trained on the full label set. }
\label{laexp}
\end{figure}

\textbf{Label sampling.}
\label{label-sampling}
To speedup computation it is possible to train our model by sampling labels, instead of training over the whole label set. How much speed-up can we achieve from this label sampling approach and still retain good levels of performance? In Figure \ref{laexp}, we attempt to answer this question by reporting the performance of our GILE-HNN model when varying the amount of labels (\%) that it uses for training over English general and specific labels of the DW dataset. In both cases, the performance of GILE-HNN tends to increase as the percentage of labels sampled increases, but it levels off for the higher percentages.

For general labels, top performance is reached with a 40\% to 50\% sampling rate, which translates to a 22\% to 18\% speedup, while for the specific labels, it is reached with a 60\% to 70\% sampling rate, which translates to a 40\% to 36\% speedup. The speedup is correlated to the size of the label set, since there are many fewer general labels than specific labels, namely 327 vs 1,058 here.  Hence, we expect even higher speedups for bigger label sets. Interestingly, GILE-HNN with label sampling reaches the performance of the  baseline with a 25\% and 60\% sample for  general and specific labels respectively. This translates to a speedup of 30\% and 50\% respectively compared to a GILE-HNN trained over all labels. Overall, these results show that our model is effective and that it can also scale to large label sets. The label sampling should also be useful in tasks where the computation resources may be limited or budgeted.

\section{Related Work}

\subsection{Neural text Classification}
Research in \textit{neural text classification} was initially based on feed-forward networks, which required unsupervised pre-training \cite{collobert11,mikolov13,le14} and later on they focused on networks with hierarchical structure. \citet{kim14} proposed a convolutional neural network (CNN) for sentence classification. \citet{johnson14} proposed a CNN for high-dimensional data classification, while \citet{zhang15} adopted a character-level CNN for text classification. \citet{lai15} proposed a recurrent CNN to capture sequential information, which outperformed simpler CNNs. \citet{lin15} and  \citet{tang15} proposed hierarchical recurrent neural networks and showed that they were superior to CNN-based models. \citet{yang16} demonstrated that a hierarchical attention network with bi-directional gated encoders outperforms previous alternatives. \citet{pappas17c} adapted such networks to learn hierarchical document structures with shared components across different languages.

The issue of scaling to large label sets has been addressed previously by output layer approximations \cite{Morin+al-2005} and with the use of sub-word units or character-level modeling \cite{sennrich15,lee16} which is mainly applicable to structured prediction problems. Despite the numerous studies, most of the existing neural text classification models ignore label descriptions and semantics. Moreover, they  are based on typical output layer parametrizations which are dependent on the label set size, and thus  are not able to scale well to large label sets nor to  generalize to unseen labels. Our output layer parametrization addresses these limitations and could potentially improve such models.

\subsection{Output Representation Learning}
There exist studies which aim to learn output representations directly from data without any semantic grounding to word embeddings \cite{Srikumar14,YehWKW17,augenstein18}. Such methods have a label-set-size dependent parametrization, which makes them data hungry,  less scalable on large label sets and incapable of generalizing to unseen classes. \citet{P18-1216} addressed the lack of semantic grounding to word embeddings by proposing an efficient method based on label-attentive text representations which are helpful for text classification. However, in contrast to our study, their parametrization is still label-set-size dependent and thus their model is not able to scale well to large label sets nor to generalize to unseen labels.

\subsection{Zero-shot Text Classification}
Several studies have focused on learning joint input-label representations grounded to word semantics for unseen label prediction for images  \cite{Weston2011,Socher13,norouzi14,ZhangGS16,fu18}, called zero-shot classification.  However, there are fewer such studies for text classification. \citet{dauphin14} predicted semantic utterances of text by mapping them in the same semantic space with the class labels using an unsupervised learning objective. \citet{yazdani15} proposed a zero-shot spoken language understanding model based on a bilinear input-label model able to generalize to previously unseen labels.  \citet{Nam16}, proposed  a bilinear joint document-label embedding which learns shared word representations between documents and labels. More recently, \citet{lei17} proposed an approach for open-world classification which aims to identify novel documents during testing but it is not able to generalize to unseen classes. Perhaps, the most similar model to ours is from the recent study by \citet{pappas18b} on neural machine translation, with the difference that they have single-word label descriptions and they use a label-set-dependent bias in a softmax linear prediction unit, which is designed for structured prediction. Hence, their model can neither handle unseen labels nor multi-label classification, as we do here.

Compared to previous joint input-label models, the proposed model has a more general and flexible parametrization which allows the output layer capacity to be controlled.
Moreover, it is not restricted to linear mappings, which have limited expressivity, but uses nonlinear mappings, similar to energy-based learning networks \cite{LeCun06atutorial,belanger16}. The link to the latter can be made if we regard $P_{val}^{(ij)}$ in Eq. \ref{prob} as an energy function for the $i$-th document and the $j$-th label, the calculation of which uses a simple multiplicative transformation (Eq. \ref{joint_space}). Lastly, the proposed model performs well on both seen and unseen label sets by leveraging the binary cross-entropy loss, which is the standard loss for classification problems, instead of a ranking loss.

\section{Conclusion}
We proposed a novel joint input-label embedding model for neural text classification which generalizes over existing input-label models and addresses their limitations while preserving high performance on both seen and unseen labels.
Compared to baseline neural models with a typical output layer, our model is more scalable and has better performance on the seen labels. Compared to previous joint input-label models, it performs significantly better on unseen labels without compromising performance on the seen labels. These improvements can be attributed to the the ability of our model to capture complex input-label relationships,  to its controllable capacity and to its training objective which is based on cross-entropy loss.

As future work, the label representation could be learned by a more sophisticated encoder, and the label sampling could benefit from importance sampling to avoid revisiting uninformative labels.  Another interesting direction would be to find a more scalable way of increasing  the  output layer capacity,  for instance using a deep rather than wide classification network.
Moreover, adapting the proposed model to structured prediction, for instance by using a softmax classification unit instead of a sigmoid one, would benefit tasks such as neural machine translation, language modeling and summarization, in isolation but also when trained jointly with multi-task learning.

\section*{Acknowledgments}

We are grateful for the support from the European Union through its Horizon 2020 program in the SUMMA project n.\ 688139, see \url{http://www.summa-project.eu}.  We would also like to thank   our action editor, Eneko Agirre, and the anonymous reviewers for their invaluable suggestions and feedback.

\bibliography{database}

\begin{thebibliography}{43}
\expandafter\ifx\csname natexlab\endcsname\relax\def\natexlab#1{#1}\fi

\bibitem[{Ammar et~al.(2016)Ammar, Mulcaire, Tsvetkov, Lample, Dyer, and
  Smith}]{ammar16}
Waleed Ammar, George Mulcaire, Yulia Tsvetkov, Guillaume Lample, Chris Dyer,
  and Noah~A. Smith. 2016.
\newblock \href {https://arxiv.org/abs/1602.01925.v2} {Massively multilingual
  word embeddings}.
\newblock \emph{CoRR}, abs/1602.01925.v2.

\bibitem[{Augenstein et~al.(2018)Augenstein, Ruder, and
  S{\o}gaard}]{augenstein18}
Isabelle Augenstein, Sebastian Ruder, and Anders S{\o}gaard. 2018.
\newblock \href {http://www.aclweb.org/anthology/N18-1172} {Multi-task learning
  of pairwise sequence classification tasks over disparate label spaces}.
\newblock In \emph{Proceedings of the 2018 Conference of the North American
  Chapter of the Association for Computational Linguistics: Human Language
  Technologies, Volume 1 (Long Papers)}, pages 1896--1906, New Orleans,
  Louisiana.

\bibitem[{Belanger and McCallum(2016)}]{belanger16}
David Belanger and Andrew McCallum. 2016.
\newblock \href {http://proceedings.mlr.press/v48/belanger16.html} {Structured
  prediction energy networks}.
\newblock In \emph{Proceedings of The 33rd International Conference on Machine
  Learning}, volume~48 of \emph{Proceedings of Machine Learning Research},
  pages 983--992, New York, New York, USA. PMLR.

\bibitem[{Chen et~al.(2015)Chen, He, Shen, Xiao, He, Gao, Song, and
  Deng}]{Sukhbaatar15}
Jianshu Chen, Ji~He, Yelong Shen, Lin Xiao, Xiaodong He, Jianfeng Gao, Xinying
  Song, and Li~Deng. 2015.
\newblock \href
  {https://papers.nips.cc/paper/5967-end-to-end-learning-of-lda-by-mirror-descent-back-propagation-over-a-deep-architecture}
  {End-to-end learning of {LDA} by mirror-descent back propagation over a deep
  architecture}.
\newblock In \emph{Advances in Neural Information Processing Systems 28}, pages
  1765--1773, Montreal, Canada.

\bibitem[{Cho et~al.(2014)Cho, van Merrienboer, Gulcehre, Bahdanau, Bougares,
  Schwenk, and Bengio}]{cho14}
Kyunghyun Cho, Bart van Merrienboer, Caglar Gulcehre, Dzmitry Bahdanau, Fethi
  Bougares, Holger Schwenk, and Yoshua Bengio. 2014.
\newblock \href
  {https://aclanthology.coli.uni-saarland.de/papers/D14-1179/d14-1179}
  {Learning phrase representations using {RNN} encoder--decoder for statistical
  machine translation}.
\newblock In \emph{Proceedings of the 2014 Conference on Empirical Methods in
  Natural Language Processing}, pages 1724--1734, Doha, Qatar.

\bibitem[{Collobert et~al.(2011)Collobert, Weston, Bottou, Karlen, Kavukcuoglu,
  and Kuksa}]{collobert11}
Ronan Collobert, Jason Weston, L{\'e}on Bottou, Michael Karlen, Koray
  Kavukcuoglu, and Pavel Kuksa. 2011.
\newblock \href {http://www.jmlr.org/papers/v12/collobert11a.html} {Natural
  language processing (almost) from scratch}.
\newblock \emph{Journal of Machine Learning Research}, 12:2493--2537.

\bibitem[{Dauphin et~al.(2014)Dauphin, T{\"{u}}r, Hakkani{-}T{\"{u}}r, and
  Heck}]{dauphin14}
Yann~N. Dauphin, G{\"{o}}khan T{\"{u}}r, Dilek Hakkani{-}T{\"{u}}r, and
  Larry~P. Heck. 2014.
\newblock \href {http://arxiv.org/abs/1401.0509} {Zero-shot learning and
  clustering for semantic utterance classification}.
\newblock In \emph{International Conference on Learning Representations},
  Banff, Canada.

\bibitem[{Firat et~al.(2016)Firat, Sankaran, Al-Onaizan, Yarman~Vural, and
  Cho}]{firat16b}
Orhan Firat, Baskaran Sankaran, Yaser Al-Onaizan, Fatos~T. Yarman~Vural, and
  Kyunghyun Cho. 2016.
\newblock \href {https://aclweb.org/anthology/D16-1026} {Zero-resource
  translation with multi-lingual neural machine translation}.
\newblock In \emph{Proceedings of the 2016 Conference on Empirical Methods in
  Natural Language Processing}, pages 268--277, Austin, USA.

\bibitem[{Frome et~al.(2013)Frome, Corrado, Shlens, Bengio, Dean, Ranzato, and
  Mikolov}]{frome13}
Andrea Frome, Greg~S. Corrado, Jon Shlens, Samy Bengio, Jeff Dean, Marc~Aurelio
  Ranzato, and Tomas Mikolov. 2013.
\newblock \href
  {http://papers.nips.cc/paper/5204-devise-a-deep-visual-semantic-embedding-model.pdf}
  {{DeViSE}: A deep visual-semantic embedding model}.
\newblock In C.~J.~C. Burges, L.~Bottou, M.~Welling, Z.~Ghahramani, and K.~Q.
  Weinberger, editors, \emph{Advances in Neural Information Processing Systems
  26}, pages 2121--2129. Curran Associates, Inc.

\bibitem[{Fu et~al.(2018)Fu, Xiang, Jiang, Xue, Sigal, and Gong}]{fu18}
Yanwei Fu, Tao Xiang, Yu-Gang Jiang, Xiangyang Xue, Leonid Sigal, and Shaogang
  Gong. 2018.
\newblock \href {https://doi.org/10.1109/MSP.2017.2763441} {Recent advances in
  zero-shot recognition: Toward data-efficient understanding of visual
  content}.
\newblock \emph{IEEE Signal Processing Magazine}, 35(1):112--125.

\bibitem[{Johnson and Zhang(2015)}]{johnson14}
Rie Johnson and Tong Zhang. 2015.
\newblock \href
  {https://aclanthology.coli.uni-saarland.de/papers/N15-1011/n15-1011}
  {Effective use of word order for text categorization with convolutional
  neural networks}.
\newblock In \emph{Proceedings of the 2015 Conference of the North American
  Chapter of the Association for Computational Linguistics: Human Language
  Technologies}, pages 103--112, Denver, Colorado.

\bibitem[{Kim(2014)}]{kim14}
Yoon Kim. 2014.
\newblock \href {https://aclanthology.info/papers/D14-1181/d14-1181}
  {Convolutional neural networks for sentence classification}.
\newblock In \emph{Proceedings of the 2014 Conference on Empirical Methods in
  Natural Language Processing}, pages 1746--1751, Doha, Qatar.

\bibitem[{Klementiev et~al.(2012)Klementiev, Titov, and
  Bhattarai}]{klementiev12}
Alexandre Klementiev, Ivan Titov, and Binod Bhattarai. 2012.
\newblock \href {http://www.aclweb.org/anthology/C12-1089} {Inducing
  crosslingual distributed representations of words}.
\newblock In \emph{Proceedings of COLING 2012}, pages 1459--1474, Mumbai,
  India.

\bibitem[{Kumar et~al.(2015)Kumar, Irsoy, Su, Bradbury, English, Pierce,
  Ondruska, Gulrajani, and Socher}]{kumar15}
Ankit Kumar, Ozan Irsoy, Jonathan Su, James Bradbury, Robert English, Brian
  Pierce, Peter Ondruska, Ishaan Gulrajani, and Richard Socher. 2015.
\newblock \href {http://proceedings.mlr.press/v48/kumar16.html} {Ask me
  anything: Dynamic memory networks for natural language processing}.
\newblock In \emph{Proceedings of The 33rd International Conference on Machine
  Learning}, pages 334--343, New York City, USA.

\bibitem[{Lai et~al.(2015)Lai, Xu, Liu, and Zhao}]{lai15}
Siwei Lai, Liheng Xu, Kang Liu, and Jun Zhao. 2015.
\newblock \href
  {https://www.aaai.org/ocs/index.php/AAAI/AAAI15/paper/view/9745/9552}
  {Recurrent convolutional neural networks for text classification}.
\newblock In \emph{Proceedings of the 29th AAAI Conference on Artificial
  Intelligence}, pages 2267--2273, Austin, USA.

\bibitem[{Le and Mikolov(2014)}]{le14}
Quoc~V. Le and Tomas Mikolov. 2014.
\newblock \href {http://proceedings.mlr.press/v32/le14.html} {Distributed
  representations of sentences and documents}.
\newblock In \emph{Proceedings of The 31st International Conference on Machine
  Learning}, pages 1188–--1196, Beijing, China.

\bibitem[{LeCun et~al.(2006)LeCun, Chopra, Hadsell, Huang, and
  et~al.}]{LeCun06atutorial}
Yann LeCun, Sumit Chopra, Raia Hadsell, Fu~Jie Huang, and et~al. 2006.
\newblock \href {http://yann.lecun.com/exdb/publis/pdf/lecun-06.pdf} {A
  tutorial on energy-based learning}.
\newblock In \emph{Predicting Structured Data}. MIT Press.

\bibitem[{Lee et~al.(2017)Lee, Cho, and Hofmann}]{lee16}
Jason Lee, Kyunghyun Cho, and Thomas Hofmann. 2017.
\newblock \href
  {https://aclanthology.coli.uni-saarland.de/papers/Q17-1026/q17-1026} {Fully
  character-level neural machine translation without explicit segmentation}.
\newblock \emph{Transactions of the Association for Computational Linguistics},
  5:365--378.

\bibitem[{Lin et~al.(2015)Lin, Liu, Yang, Li, Zhou, and Li}]{lin15}
Rui Lin, Shujie Liu, Muyun Yang, Mu~Li, Ming Zhou, and Sheng Li. 2015.
\newblock \href
  {https://aclanthology.coli.uni-saarland.de/papers/D15-1106/d15-1106}
  {Hierarchical recurrent neural network for document modeling}.
\newblock In \emph{Proceedings of the 2015 Conference on Empirical Methods in
  Natural Language Processing}, pages 899--907, Lisbon, Portugal.

\bibitem[{Luong et~al.(2015)Luong, Pham, and Manning}]{thang15}
Thang Luong, Hieu Pham, and Christopher~D. Manning. 2015.
\newblock \href {https://aclanthology.info/papers/D15-1166/d15-1166} {Effective
  approaches to attention-based neural machine translation}.
\newblock In \emph{Proceedings of the 2015 Conference on Empirical Methods in
  Natural Language Processing}, pages 1412--1421, Lisbon, Portugal.

\bibitem[{Mensink et~al.(2012)Mensink, Verbeek, Perronnin, and
  Csurka}]{mensik12}
Thomas Mensink, Jakob Verbeek, Florent Perronnin, and Gabriela Csurka. 2012.
\newblock Metric learning for large scale image classification: Generalizing to
  new classes at near-zero cost.
\newblock In \emph{Computer Vision -- ECCV 2012}, pages 488--501, Berlin,
  Heidelberg. Springer Berlin Heidelberg.

\bibitem[{Mikolov et~al.(2013)Mikolov, Sutskever, Chen, Corrado, and
  Dean}]{mikolov13}
Tomas Mikolov, Ilya Sutskever, Kai Chen, Greg~S Corrado, and Jeff Dean. 2013.
\newblock \href
  {https://papers.nips.cc/paper/5021-distributed-representations-of-words-and-phrases-and-their-compositionality}
  {Distributed representations of words and phrases and their
  compositionality}.
\newblock In C.~J.~C. Burges, L.~Bottou, M.~Welling, Z.~Ghahramani, and K.~Q.
  Weinberger, editors, \emph{Advances in Neural Information Processing Systems
  26}, pages 3111--3119. Curran Associates, Inc.

\bibitem[{Morin and Bengio(2005)}]{Morin+al-2005}
Frederic Morin and Yoshua Bengio. 2005.
\newblock \href
  {https://www.iro.umontreal.ca/~lisa/pointeurs/hierarchical-nnlm-aistats05.pdf}
  {Hierarchical probabilistic neural network language model}.
\newblock In \emph{Proceedings of the Tenth International Workshop on
  Artificial Intelligence and Statistics}, pages 246--252.

\bibitem[{Mrini et~al.(2017)Mrini, Pappas, and Popescu-Belis}]{mrini17}
Khalil Mrini, Nikolaos Pappas, and Andrei Popescu-Belis. 2017.
\newblock \href {https://publidiap.idiap.ch/index.php/publications/show/3642}
  {Cross-lingual transfer for news article labeling: Benchmarking statistical
  and neural models}.
\newblock In \emph{Idiap Research Report}, Idiap-RR-26-2017.

\bibitem[{Nam et~al.(2016)Nam, Menc\'{\i}a, and F\"{u}rnkranz}]{Nam16}
Jinseok Nam, Eneldo~Loza Menc\'{\i}a, and Johannes F\"{u}rnkranz. 2016.
\newblock \href
  {https://www.aaai.org/ocs/index.php/AAAI/AAAI16/paper/view/12058} {All-in
  text: Learning document, label, and word representations jointly}.
\newblock In \emph{Proceedings of the 13th AAAI Conference on Artificial
  Intelligence}, AAAI'16, pages 1948--1954, Phoenix, Arizona.

\bibitem[{Norouzi et~al.(2014)Norouzi, Mikolov, Bengio, Singer, Shlens, Frome,
  Corrado, and Dean}]{norouzi14}
Mohammad Norouzi, Tomas Mikolov, Samy Bengio, Yoram Singer, Jonathon Shlens,
  Andrea Frome, Greg Corrado, and Jeffrey Dean. 2014.
\newblock \href {https://arxiv.org/abs/1312.5650} {Zero-shot learning by convex
  combination of semantic embeddings.}
\newblock In \emph{International Conference on Learning Representations},
  Banff, Canada.

\bibitem[{Pang and Lee(2005)}]{Pang2005}
Bo~Pang and Lillian Lee. 2005.
\newblock \href
  {https://aclanthology.coli.uni-saarland.de/papers/P05-1015/p05-1015} {Seeing
  stars: Exploiting class relationships for sentiment categorization with
  respect to rating scales}.
\newblock In \emph{Proceedings of the 43rd Annual Meeting on Association for
  Computational Linguistics}, pages 115--124, Ann Arbor, Michigan.

\bibitem[{Pappas et~al.(2018)Pappas, Miculicich, and Henderson}]{pappas18b}
Nikolaos Pappas, Lesly Miculicich, and James Henderson. 2018.
\newblock \href {http://www.aclweb.org/anthology/W18-6308} {Beyond weight
  tying: Learning joint input-output embeddings for neural machine
  translation}.
\newblock In \emph{Proceedings of the Third Conference on Machine Translation:
  Research Papers}, pages 73--83, Belgium, Brussels. Association for
  Computational Linguistics.

\bibitem[{Pappas and Popescu-Belis(2017)}]{pappas17c}
Nikolaos Pappas and Andrei Popescu-Belis. 2017.
\newblock \href {http://aclweb.org/anthology/I17-1102} {Multilingual
  hierarchical attention networks for document classification}.
\newblock In \emph{Proceedings of the Eighth International Joint Conference on
  Natural Language Processing (Volume 1: Long Papers)}, pages 1015--1025.

\bibitem[{Rush et~al.(2015)Rush, Chopra, and Weston}]{rush15}
Alexander~M. Rush, Sumit Chopra, and Jason Weston. 2015.
\newblock \href
  {https://aclanthology.coli.uni-saarland.de/papers/D15-1044/d15-1044} {A
  neural attention model for abstractive sentence summarization}.
\newblock In \emph{Proceedings of the 2015 Conference on Empirical Methods in
  Natural Language Processing}, pages 379--389, Lisbon, Portugal.

\bibitem[{Sennrich et~al.(2016)Sennrich, Haddow, and Birch}]{sennrich15}
Rico Sennrich, Barry Haddow, and Alexandra Birch. 2016.
\newblock \href
  {https://aclanthology.coli.uni-saarland.de/papers/P16-1162/p16-1162} {Neural
  machine translation of rare words with subword units}.
\newblock In \emph{Proceedings of the 54th Annual Meeting of the Association
  for Computational Linguistics (Volume 1: Long Papers)}, pages 1715--1725,
  Berlin, Germany.

\bibitem[{Shu et~al.(2017)Shu, Xu, and Liu}]{lei17}
Lei Shu, Hu~Xu, and Bing Liu. 2017.
\newblock \href {https://www.aclweb.org/anthology/D17-1314} {{DOC}: Deep open
  classification of text documents}.
\newblock In \emph{Proceedings of the 2017 Conference on Empirical Methods in
  Natural Language Processing}, pages 2911--2916, Copenhagen, Denmark.
  Association for Computational Linguistics.

\bibitem[{Socher et~al.(2013)Socher, Ganjoo, Manning, and Ng}]{Socher13}
Richard Socher, Milind Ganjoo, Christopher~D. Manning, and Andrew~Y. Ng. 2013.
\newblock \href
  {https://papers.nips.cc/paper/5027-zero-shot-learning-through-cross-modal-transfer}
  {Zero-shot learning through cross-modal transfer}.
\newblock In \emph{Proceedings of the 26th International Conference on Neural
  Information Processing Systems}, NIPS'13, pages 935--943, Lake Tahoe, Nevada.

\bibitem[{Srikumar and Manning(2014)}]{Srikumar14}
Vivek Srikumar and Christopher~D. Manning. 2014.
\newblock \href {http://dl.acm.org/citation.cfm?id=2969033.2969191} {Learning
  distributed representations for structured output prediction}.
\newblock In \emph{Proceedings of the 27th International Conference on Neural
  Information Processing Systems - Volume 2}, NIPS'14, pages 3266--3274,
  Cambridge, MA, USA. MIT Press.

\bibitem[{Tang et~al.(2015)Tang, Qin, and Liu}]{tang15}
Duyu Tang, Bing Qin, and Ting Liu. 2015.
\newblock \href {https://doi.org/10.18653/v1/D15-1167} {Document modeling with
  gated recurrent neural network for sentiment classification}.
\newblock In \emph{Proceedings of the 2015 Conference on Empirical Methods in
  Natural Language Processing}, pages 1422--1432, Lisbon, Portugal. Association
  for Computational Linguistics.

\bibitem[{Wang et~al.(2018)Wang, Li, Wang, Zhang, Shen, Zhang, Henao, and
  Carin}]{P18-1216}
Guoyin Wang, Chunyuan Li, Wenlin Wang, Yizhe Zhang, Dinghan Shen, Xinyuan
  Zhang, Ricardo Henao, and Lawrence Carin. 2018.
\newblock \href {http://aclweb.org/anthology/P18-1216} {Joint embedding of
  words and labels for text classification}.
\newblock In \emph{Proceedings of the 56th Annual Meeting of the Association
  for Computational Linguistics (Volume 1: Long Papers)}, pages 2321--2331.
  Association for Computational Linguistics.

\bibitem[{Weston et~al.(2010)Weston, Bengio, and Usunier}]{Weston10}
Jason Weston, Samy Bengio, and Nicolas Usunier. 2010.
\newblock \href {https://doi.org/10.1007/s10994-010-5198-3} {Large scale image
  annotation: Learning to rank with joint word-image embeddings}.
\newblock \emph{Mach. Learn.}, 81(1):21--35.

\bibitem[{Weston et~al.(2011)Weston, Bengio, and Usunier}]{Weston2011}
Jason Weston, Samy Bengio, and Nicolas Usunier. 2011.
\newblock \href {https://ai.google/research/pubs/pub37180} {{WSABIE}: Scaling
  up to large vocabulary image annotation}.
\newblock In \emph{Proceedings of the Twenty-Second International Joint
  Conference on Artificial Intelligence (Volume 3)}, pages 2764--2770,
  Barcelona, Spain.

\bibitem[{Yang et~al.(2016)Yang, Yang, Dyer, He, Smola, and Hovy}]{yang16}
Zichao Yang, Diyi Yang, Chris Dyer, Xiaodong He, Alex Smola, and Eduard Hovy.
  2016.
\newblock \href
  {https://aclanthology.coli.uni-saarland.de/papers/N16-1174/n16-1174}
  {Hierarchical attention networks for document classification}.
\newblock In \emph{Proceedings of the 2016 Conference of the North American
  Chapter of the Association for Computational Linguistics: Human Language
  Technologies}, pages 1480--1489, San Diego, California.

\bibitem[{Yazdani and Henderson(2015)}]{yazdani15}
Majid Yazdani and James Henderson. 2015.
\newblock \href
  {https://aclanthology.coli.uni-saarland.de/papers/D15-1027/d15-1027} {A model
  of zero-shot learning of spoken language understanding}.
\newblock In \emph{Proceedings of the 2015 Conference on Empirical Methods in
  Natural Language Processing}, pages 244--249, Lisbon, Portugal.

\bibitem[{Yeh et~al.(2018)Yeh, Wu, Ko, and Wang}]{YehWKW17}
Chih{-}Kuan Yeh, Wei{-}Chieh Wu, Wei{-}Jen Ko, and Yu{-}Chiang~Frank Wang.
  2018.
\newblock \href {https://arxiv.org/abs/1707.00418} {Learning deep latent spaces
  for multi-label classification}.
\newblock In \emph{In Proceedings of the 32nd AAAI Conference on Artificial
  Intelligence}, New Orleans, USA.

\bibitem[{Zhang et~al.(2015)Zhang, Zhao, and LeCun}]{zhang15}
Xiang Zhang, Junbo Zhao, and Yann LeCun. 2015.
\newblock \href
  {http://papers.nips.cc/paper/5782-character-level-convolutional-networks-for-text-classification.html}
  {Character-level convolutional networks for text classification}.
\newblock In \emph{Advances in Neural Information Processing Systems 28}, pages
  649--657, Montreal, Canada.

\bibitem[{Zhang et~al.(2016)Zhang, Gong, and Shah}]{ZhangGS16}
Yang Zhang, Boqing Gong, and Mubarak Shah. 2016.
\newblock \href {https://arxiv.org/abs/1605.09759} {Fast zero-shot image
  tagging}.
\newblock In \emph{Proceedings of the IEEE Conference on Computer Vision and
  Pattern Recognition}, Las Vegas, USA.

\end{thebibliography}
\bibliographystyle{acl_natbib}

\end{document}